\PassOptionsToPackage{table}{xcolor}
\documentclass[10pt,twocolumn,letterpaper]{article}

\usepackage[pagenumbers]{cvpr}              %

\usepackage[english]{babel}

\usepackage{amsfonts}
\usepackage{amsmath}
\usepackage{amssymb}
\usepackage{amsthm}

\usepackage{thm-restate} %

\usepackage{algorithm}
\usepackage{algpseudocode}

\usepackage{multirow}
\usepackage{graphicx}
\usepackage{amsmath}

\definecolor{myrowcolor}{rgb}{0.9, 0.95, 1}
\definecolor{mygreen}{RGB}{144, 238, 144}   %
\definecolor{myorange}{RGB}{255, 200, 124}  %
\definecolor{myred}{RGB}{255, 153, 153}     %

\newcommand{\trim}[1]{\hspace{-#1pt}}
\newcommand{\shorteq}{\hspace{-3pt}=\hspace{-3pt}}
\newcommand{\short}[2]{\hspace{-#2pt}#1\hspace{-#2pt}}

\newcommand\blfootnote[1]{%
  \begingroup
  \renewcommand\thefootnote{}\footnote{#1}%
  \addtocounter{footnote}{-1}%
  \endgroup
}

\usepackage{layouts}

\newcommand{\Real}{\mathbb{R}}
\newcommand{\X}{\mathcal{X}}
\newcommand{\Y}{\mathcal{Y}}
\newcommand{\M}{\mathcal{M}}
\newcommand{\Sphere}{\mathbb{S}}
\newcommand{\Z}{\mathcal{Z}}
\newcommand{\attrs}{\Z_{attr}}
\newcommand{\objs}{\Z_{obj}}
\newcommand{\T}{\mathcal{T}}
\newcommand{\Zextra}{\mathcal{E}}
\newcommand{\Zslice}{\Z(z_i)}
\newcommand{\Prob}{\mathbb{P}}

\newcommand{\uu}{\mathbf{u}}
\newcommand{\vv}{\mathbf{v}}

\newcommand{\vz}[1]{\vv_{z_{#1}}}
\newcommand{\ut}{\Tilde{\uu}}

\newcommand{\scorez}{p_{(z,e)}}
\newcommand{\encoder}{\phi}
\newcommand{\suchthat}{\, | \,}

\newcommand{\ours}{\textsc{GDE }}
\newcommand{\GDE}{\textsc{GDE }}

\newcommand{\LDE}{\textsc{LDE }}

\DeclareMathOperator*{\argmin}{arg\,min}
\DeclareMathOperator*{\argmax}{arg\,max}
\DeclareMathOperator{\Exp}{Exp}
\DeclareMathOperator{\Log}{Log}

\newcommand{\p}{\mu}
\newcommand{\q}{\uu}
\newcommand{\s}{\uu'}
\newcommand{\vvv}{\vv}

\newtheorem{prop}{Proposition}

\theoremstyle{definition}
\newtheorem{defi}{Definition}

\usepackage[table]{xcolor}

\definecolor{cvprblue}{rgb}{0.21,0.49,0.74}
\usepackage[pagebackref,breaklinks,colorlinks,allcolors=cvprblue]{hyperref}

\def\paperID{16663} %
\def\confName{CVPR}
\def\confYear{2025}

\title{Not Only Text: Exploring Compositionality of Visual Representations\\in Vision-Language Models}

\author{Davide Berasi\textsuperscript{$1$} \quad Matteo Farina\textsuperscript{$2$}\quad Massimiliano Mancini\textsuperscript{$2$}\\
Elisa Ricci\textsuperscript{$1, 2$}\quad Nicola Strisciuglio\textsuperscript{$3$}\\
\small
$^1$Fondazione Bruno Kessler \quad $^2$University of Trento \quad $^3$University of Twente
}

\begin{document}

\maketitle
\begin{abstract}
Vision-Language Models (VLMs) learn a shared feature space for text and images, enabling the comparison of inputs of different modalities. 
While prior works demonstrated that VLMs organize natural language representations into regular structures encoding composite meanings, it remains unclear if compositional patterns also emerge in the visual embedding space.
In this work, we investigate compositionality in the image domain, where the analysis of compositional properties is challenged by noise and sparsity of visual data.
We address these problems and propose a framework, called Geodesically Decomposable Embeddings (GDE), that approximates image representations with geometry-aware compositional structures in the latent space.
We demonstrate that visual embeddings of pre-trained VLMs exhibit a compositional arrangement, and evaluate the effectiveness of this property in the tasks of compositional classification and group robustness.  %
GDE achieves stronger performance in compositional classification compared to its counterpart method that assumes linear geometry of the latent space. 
Notably, it is particularly effective for group robustness, where we achieve higher results than task-specific solutions.
Our results indicate that VLMs can automatically develop a human-like form of compositional reasoning in the visual domain, making their underlying processes more interpretable. %
Code is available at {\small \url{https://github.com/BerasiDavide/vlm_image_compositionality}.}
\blfootnote{Corresponding author: {\tt dberasi@fbk.eu}.}
\end{abstract}
 
\section{Introduction}

\begin{figure}[!t]
    \centering
    \includegraphics[trim={0 0 0 15}, clip, width=\linewidth]{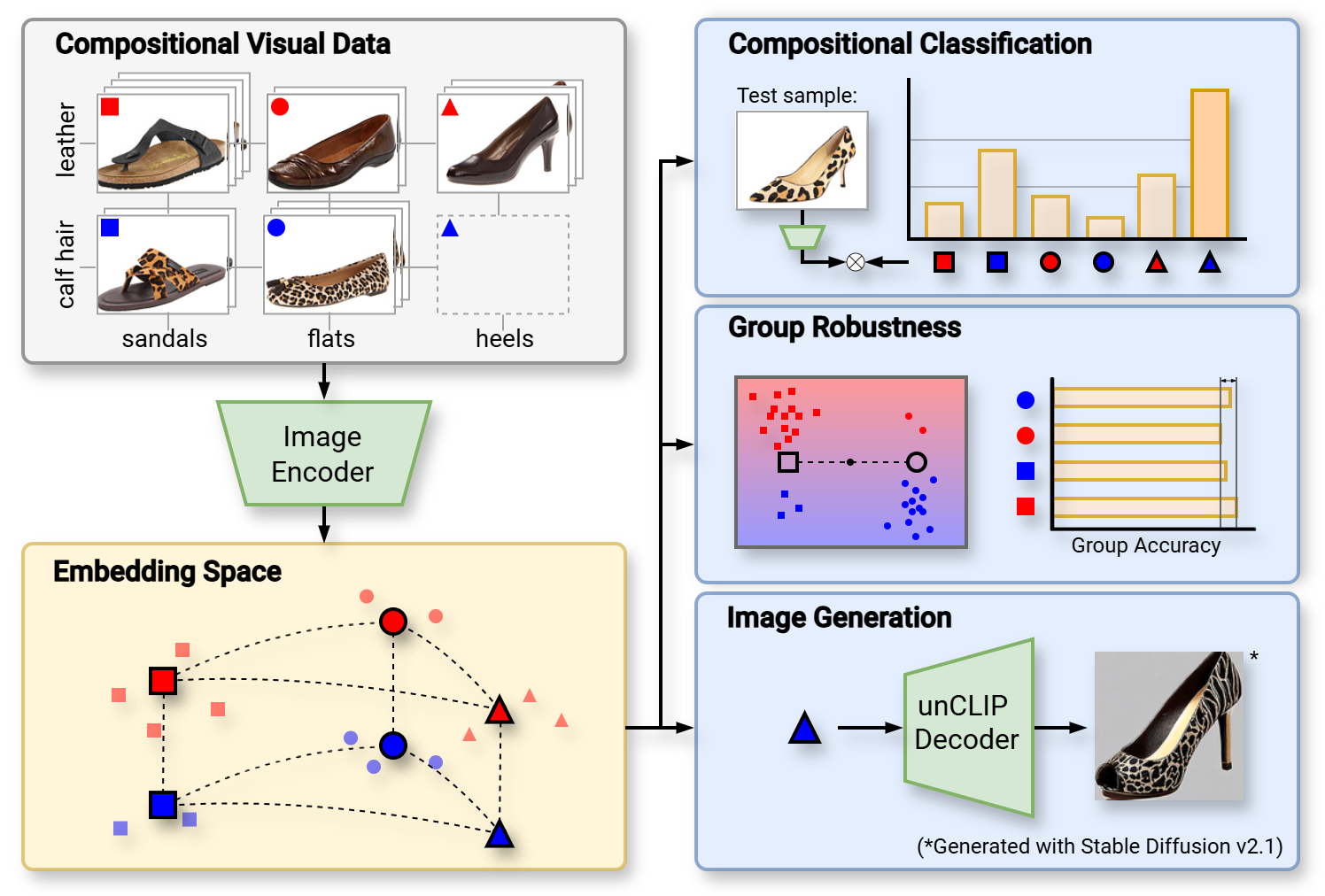}
    \caption{
    \textbf{Compositional structures in visual embedding space}.
    (\textit{left}) Pre-trained VLM represents visual inputs of composite meanings in regular geometric shapes. The modularity of these structures enables the separation of the primitive components and the composition of unseen combinations.
    (\textit{right}) We evaluate the usefulness of these properties in compositional classification, group robustness, and image generation.
    }
    \label{fig:teaser}
\end{figure}

\label{sec:intro}
Compositionality is the principle by which cognitive and computational systems create meaning of a complex expression by combining the meaning of its (simpler) parts~\cite{partee1995lexical, partee2008compositionality}. Humans leverage compositionality instinctively, combining known elements to interpret novel situations. In machine intelligence, efforts were made to replicate this capability by developing models that imitate compositional processes, \eg, solving complex tasks via sub-goals \cite{schmidhuber1990towards,papadopoulos2019make,koushik2017compositional,czechowski2021subgoal}, modeling objects as compositions of their parts \cite{fischler1973representation,felzenszwalb2008discriminatively,saralajew2019classification,si2013learning,ommer2009learning}, encoding concept hierachies \cite{desai2023hyperbolic, ge2023hyperbolic, ramasinghe2024accept, kusupati2022matryoshka}, explicitly learning compositional representations~\cite{andreas2019measuring,liao2016learning,gong2018learning,nagarajan2018attributes}, or architectures ~\cite{hudson2018compositional,wang2015semantic,stone2017teaching,li2019aognets,higgins2018scan}.

With the rise of modern Vision-Language Models (VLMs) \cite{radford2021learning, jia2021scaling, zhang2021explainable} jointly trained on large-scale image-text pairs, there has been growing interest in investigating whether these models exhibit instrinsic compositional behaviors~\cite{yun2023do,perera2023prompt, nayak2023learning}.
In particular, Trager \textit{et al.}~\cite{trager2023linear} investigated latent compositional structures within the CLIP~\cite{radford2021learning} %
text embedding space, demonstrating that composite concepts can be represented as linear combinations of embedding vectors corresponding to various factors. These vectors, called \textit{ideal words}, can be used to compose new concepts in the embedding space. 
Their work focuses on finding compositional structures in the text embedding space of CLIP, motivated by the fact that the structured and symbolic nature of language may facilitate the study of computational approaches to capture compositional meaning. 
However, cognitive studies show that language itself is used to describe and interpret the visual world and directly affects visual perception~\cite{Chabal2015language}. Hence, similar to text, human visual representations exhibit a compositional structure~\cite{HafriGreenFirestone2023}, made of simpler components systematically combined. 
Despite this connection, compositional properties of visual embeddings of VLMs have remained so far mostly unexplored.

To fill this gap, in this paper, we introduce \textsc{Geodesically Decomposable Embeddings} (\textsc{GDE}), a framework grounded in differential geometry and designed to investigate compositional structures of pre-trained embeddings within Riemannian manifolds (Fig.\ref{fig:teaser}). %
Visual embeddings exhibit unique challenges not present in compositional analysis of text embeddings, namely \emph{data sparsity} in the compositional space and \emph{noise and ambiguity} in images. 
Specifically, we deal with the sparsity of composite concepts, as certain combinations of elementary primitives may not appear in real image collections (\textit{e.g.}, focusing on objects and attributes, ``blue dog'' images are unlikely to exist). %
Noise and ambiguity concern additional visual cues and information present in images, e.g. background, context, etc., that do not correspond to the composite concepts. 
We evaluated the compositional representations computed with the proposed approach in two relevant applications, namely compositional classification and group robustness (Fig. \ref{fig:teaser}), considering publicly available datasets, showing that it better captures visual compositional structures than the alternatives (\eg,~\cite{dehdashtian2024fairerclip}). GDE is particularly effective for group robustness, where we achieve better debiasing results than task-specific methods. 
Furthermore, we show that \ours can be successfully used in combination with state-of-the-art generative models to synthesize images of compositional concepts.
Our contributions are: 
\begin{enumerate}[i)]
    \item We study compositional structures within visual embeddings for VLMs and demonstrate that the latent representations of visual signals also exhibit a degree of compositionality.
    \item We show that, unlike for text embeddings, linear structures are insufficient to (de)compose visual concepts; thus, the manifold geometry must be considered.
    \item We propose a framework that deals with the sparsity and noise of composite concepts in images, enabling the compositional analysis of visual embeddings.
\end{enumerate}

\section{Related Work}

\noindent\textbf{Compositionality in Vision.} Compositionality is considered a cornerstone of perception~\cite{Lande2024perception}, and compositional representations offer an effective tool to represent real-world phenomena~\cite{Feldman2024probabilistic}. The primary benefit of compositionality is the possibility of combining the representation of simpler concepts to understand and reason on complex ones, allowing for generalization to new unseen combinations of concepts~\cite{saralajew2019classification,nagarajan2018attributes,koushik2017compositional}. In computer vision, early efforts focused on recognizing objects as composition of parts~\cite{ommer2007learning,ommer2009learning,felzenszwalb2008discriminatively,fischler1973representation} and evolved into architectures that can recognize and model objects in a compositional fashion~\cite{saralajew2019classification}, compositional generation~\cite{papadopoulos2019make,tan2019text2scene,zhao2018modular}, and interpretable representations~\cite{stone2017teaching,bohle2022b}. Compositionality has also lead to progress in various tasks, such as %
human-object interaction detection, model spatial/semantic relationships~\cite{kato2018compositionalhoi,hou2020visualhoi,hou2022discoveringhoi}, and compositional zero-shot learning, where the goal is to recognize unseen compositions of training primitives~\cite{mancini2021open,mancini2022learning,nagarajan2018attributes,li2020symmetry,misra2017red}. %
While these works focus on specific applications, in this paper we aim to study whether there exists an underlying compositional structure in the visual embeddings of VLMs. 

\noindent\textbf{Compositionality in VLMs.}
Modern Vision-Language Models (VLMs) like CLIP~\cite{radford2021learning} are trained to extract meaningful representations from complex visual scenes guided by textual inputs without a priori imposing any form of compositionality. In this context, a natural question is: \textit{Does compositional behavior emerge automatically in VLMs?}
 
Previous works already showed how VLMs are more suitable for tasks such as compositional zero-shot learning~\cite{perera2023prompt,nayak2023learning,lu2023decomposed}, and how their representations allow for cross-modal compositions, such as visual editing~\cite{kawar2023imagic,zhang2023sine,ceylan2023pix2video} and compositional retrieval~\cite{saito2023pic2word,baldrati2023zero,karthik2024vision,hsieh2024sugarcrepe}. At the same time, works studied the challenges of VLMs in model compositional inputs, \eg, at the level of word order, object-attribute bindings, spatial relationships and other compositional challenges~\cite{yuksekgonul2022and,thrush2022winoground,hsieh2024sugarcrepe,tong2024eyes}.

In this paper, we %
study the compositional structure in the visual embeddings extracted from VLMs. Close to our goal is %
~\cite{Lewis2024bind}, studying the compositional properties of the CLIP text encoder %
through compositional distributional semantics models in synthetic test scenarios. %
Similarly, \cite{trager2023linear} show that the textual embeddings of VLMs %
can be well approximated by linear compositions of smaller sets of ideal vectors. 
Motivated by the cross-modal alignment of VLMs, we investigate whether the embeddings %
of visual inputs exhibit an analogous compositional property. 
We achieve this by constructing a geometry-aware decomposition framework, following ideas similar to \cite{oldfield2023parts}, where Principal Geodesic Analysis (PGA) \cite{fletcher2004principal} is applied to learn lower-dimensional submanifolds of the CLIP sphere that are associated to distinct parts-of-speech. 
To the best of our knowledge, this is the first work that investigates the emergence of compositional structures in the visual embeddings of VLMs.

\section{Method}
\label{sec:method}

We propose a framework to analyze the compositional properties of image embeddings of neural encoders. We start by reviewing the fundamentals of the CLIP model along with key concepts from differential geometry (\cref{sec:preliminaries}).
We then formalize the concept of geodesic decomposability (\cref{sec:geodesically dec. embs}) and we discuss our methodology for dealing with visual inputs (\cref{sec:GDE from visual inputs}).

\subsection{Preliminaries}
\label{sec:preliminaries}

\noindent\textbf{Contrastive Language-Image Pretraining (CLIP)} consists of a pre-trained image encoder $\phi_{im}:\X \to \Real^d$ and a text encoder $\phi_{t}:\Y \to \Real^d$ that represent multi-modal text-visual inputs in a shared vision-language space.
The latent representations of an image $x \in \X$ and text $y \in \Y$ are compared by cosine similarity, which is the scalar product $\uu_x^\top \uu_y$ of their normalized versions $\uu_x = \phi_{im}(x)/||\phi_{im}(x)||$, $\uu_y = \phi_t(y)/||\phi_t(y)||$.
The weights of the encoders are trained to optimize a contrastive objective on a huge collection of paired image-text samples.
Since the norm of CLIP embeddings does not carry any meaningful information, spherical geometry applies to their post-hoc analysis.

\noindent\textbf{Riemannian Manifolds} 
are geometric spaces where intrinsic distances can be measured.
For a generic manifold $\M \subset \Real^d$ with intrinsic distance $d_{\M}:\M\times\M \to [0, \infty)$, we now recall the notions of \textit{exponential map} and \textit{intrinsic mean}. 
These tools permit operating with non-linear data, like the spherical normalized CLIP embeddings, while respecting their intrinsic shape.
Let $\p$ be a point on $\M$ and let $T_\p\M$ be the tangent space in $\p$. The \textit{exponential map} projects a tangent vector $\vvv \in T_\p\M$ onto the manifold by moving along the geodesic segment it defines. Formally, if $\gamma_\vvv:[0, 1] \to \M$ is the unique geodesic path starting from $\gamma_\vvv(0) = \p$ with initial velocity $\dot\gamma_\vvv(0) = \vvv$, then $\Exp_\mu(\vvv) := \gamma_\vvv(1)$.
This function is locally invertible and its inverse is the logarithmic map $\Log_\p = \Exp_\p^{-1}$.
The exponential and logarithmic maps send straight lines of the tangent plane into geodesic curves of the manifold, and vice-versa. 
Moreover, they approximately preserve distances between elements close to the point of tangency $\p$:
\begin{equation}\label{eq:log is approx isometry}
    d_\M(\q,\s) \approx ||\Log_\p(\q) - \Log_\p(\s)||, \quad \q,\s \in \M
\end{equation}
Note that in (\ref{eq:log is approx isometry}) the equality holds if $\q=\p$ or $\s=\p$.
When applying the logarithmic map to a set of points $\{\uu_i\}_{i=1}^N \subset \M$, the natural choice for the point of tangency $\mu$ is the \textit{intrinsic mean}, \ie, the element of $\M$ minimizing the average squared distance to the given points.
In a more general definition, each point $\uu_i$ $(i=1, \dots, N)$ is associated to a scalar weight $w_i$ belonging to a probability-simplex vector $\Delta_N$ 
and the (weighted) intrinsic mean is:
\begin{equation}\label{eq:intrinsic mean}
    \mu = \argmin_{\uu \in \M} \sum_{i=1}^N w_i \; d_\M(\uu, \uu_i)^2
\end{equation}
This distance-minimizing element $\mu$ guarantees that the images of the points through the logarithmic map are centered in the origin of the tangent space:
$\sum_i w_i \Log_\mu(\uu_i)=0$.

\subsection{Geodesically Decomposable Embeddings}
\label{sec:geodesically dec. embs}
We now formalize our proposed notion of compositional embeddings.
We consider a set of composite meanings $\Z = \Z_1\times\dots\times\Z_s$, defined as the Cartesian product between finite lists of primitive concepts, and refer to the $\Z_i$ $(i=1, \dots, s)$ as the \textit{dimensions} of $\Z$. %
For example, $\Z=\{\mbox{red, blue}\}\times\{\mbox{car, dress, flower}\}$ combines primitives from an attribute dimension and an object dimension.

\begin{figure}[!t]
    \centering
    \includegraphics[trim={0 0 0 0} , clip,width=\linewidth]{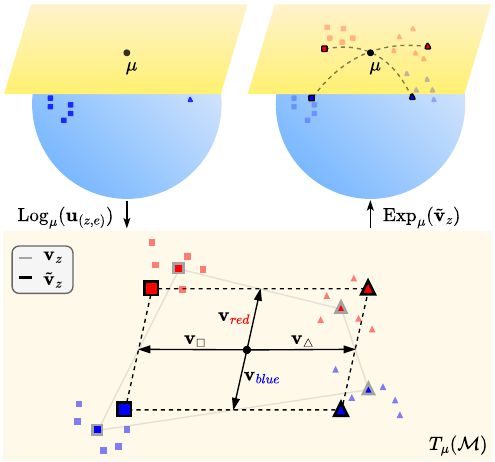}
    \caption{\textbf{Sketch of our decomposition method}. %
    (\textit{top-left}) Each concept in $\Z \trim{2}=\trim{2} \{{\color{red} red}, {\color{blue} blue}\}\times\{\Box, \triangle\}$ is represented by $k=5$ embeddings on a manifold. (\textit{bottom}) These are mapped in the tangent space where optimal primitive directions are computed as vector means and combined by addition. (\textit{top-right}) The obtained compositions are mapped back to the manifold to obtain a decomposable approximation of the input embeddings.}
    \label{fig:diagram method}
\end{figure}

We then consider an embedding map $\encoder: \Z \to \M$ representing the composite concepts as points on a manifold $\M \subset \Real^d$.
Intuitively, the set $\encoder(\Z) = \{\uu_z \suchthat z \in \Z\}$ is compositional if it has a regular structure reflecting the composite nature of the inputs, \ie, if one can \emph{compose} primitive concepts within the geometric space to obtain embeddings of complex meanings. 
In this paper, we associate compositionality to the notion of \emph{geodesic decomposability} which accounts for the intrinsic geometry of the manifold.
\begin{defi}[Geodesically decomposable embeddings]
\label{def:geodesically dec. embs}
    A set of embeddings $\encoder(\Z)=\{\uu_z \suchthat z \in \Z\} \subset \M$ with intrinsic mean $\mu$ is %
    \textit{geodesically decomposable} if there exist $\vz{i} \in T_\mu\M$ for all $z_i \in \Z_i$ $(i = 1,..., s)$ such that
    \begin{equation}
    \label{eq:uz decomposition}
            \uu_z = \Exp_\mu(\vz{1} + \dots + \vz{s}) \hspace{0.5cm} \forall z=(z_1, \dots, z_s)
    \end{equation}
\end{defi}
Note that in a decomposable set $\phi(\Z)$ a new valid decomposition is obtained by adding the same tangent vector to all $\vz{i}$ and subtracting it from all $\vz{j}$, for any $i \ne j$. However, we can guarantee the uniqueness of the factorization by imposing a centering constraint.
\begin{restatable}{lemma}{lemmauniqueness}
\label{lemma:decomposition is unique}
    Let $\encoder(\Z)$ be a geodesically decomposable set. Then there exist unique vectors $\vz{i} \in T_\mu\M$ for all $z_i \in \Z_i$ %
    such that $\sum_{z_i \in \Z_i} \vz{i} = 0$ for all $i=1, \dots, s$ and \cref{eq:uz decomposition} holds.
\end{restatable}
\noindent%
For an intuitive interpretation, the intrinsic mean $\mu$ of a decomposable set can be seen as the \textit{context} of the decomposition, and each unique direction $\vz{i}$ represents the meaning of the primitive concept $z_i$ relative to $\mu$. 
These ``universal directions'' are combined by addition on the tangent space $T_\mu\M$.
The exponential map of the resulting tangent vector defines the geodesic segment on the manifold $\M$ from $\mu$ to the corresponding composite meaning (see~\cref{fig:diagram method}).

Our notion of geodesic decomposability is general and applicable to manifolds of any shape.
It generalizes that of~\cite{trager2023linear}, which is equivalent to ours in the special case $\M=\Real^n$, where
the intrinsic mean is the arithmetic mean, and the exponential and logarithmic maps behave like the identity function. 
Our manifold formalization agrees with the fact that lower-dimensional semantic subspaces in CLIP latent space are captured by submanifolds better than linear subspaces~\cite{oldfield2023parts}.

\noindent\textbf{Best decomposable approximation.}
Decomposable sets live in a lower dimension subspace of their manifold $\M$.
The dimension of $Span(\{\vz{i}\}_{z_i\in\Z_i})$ is indeed at most $|\Z_i|-1$ for all $i=1, \dots, s$, implying the additive combinations of the primitive directions belong to a subspace of dimension %
at most $\sum_i(|\Z_i|-1)$.
This suggests that a generic set of embeddings $\{\uu_z\}$ is unlikely to be perfectly %
decomposable.
We thus search for its best decomposable approximation, 
that is the set $\{\ut_z\}$ that minimizes the error
\begin{equation}\label{eq:geodesic objective (simple)}
    \sum_{z \in \Z} d_\M(\uu_z, \ut_z)^2
\end{equation}
In general, this is a hard problem to solve. %
Similarly to the standard solution to Principal Geodesic Analysis \cite{fletcher2004principal}, we use~\cref{eq:log is approx isometry} to approximate the %
objective in the ``simpler'' Euclidean space $T_\mu\M$,
and rewrite~\cref{eq:geodesic objective (simple)} as:
\begin{equation}\label{eq:linear objective}
    \sum_{z \in \Z} ||\Log_\mu(\uu_{z}) - \Log_\mu(\ut_z) ||^2,
\end{equation}
\noindent The solution to the approximate problem is obtained by computing vector means in $T_\mu\M$, as described in the next proposition. 
For a fixed primitive concept $z_i \in \Z_i$, let
$
    \Zslice = \{(z_1', \dots, z_r') \in \Z \suchthat z_i'=z_i\}
$
denote the \emph{slice} of $\Z$ containing all tuples with the $i$-th component equal to $z_i$.
\begin{prop}
\label{prop:best decomposable approx (simple)}
Given a set $\encoder(\Z)=\{\uu_z \suchthat z \in \Z\} \subset \M$ with intrinsic mean $\mu$, the minimization problem
\begin{equation}
\begin{aligned}
    &\argmin_{\{\ut_z\}} \sum_{z \in \Z} ||\Log_\mu(\uu_z) - \Log_\mu(\ut_z)||^2, \\ 
    &\qquad s.t. \; \{\ut_z\} \mbox{ is geodesically decomposable}
\end{aligned}
\end{equation}
is solved by $\ut_z = \Exp_\mu(\vz{1} + \dots + \vz{r})$, where
\begin{equation}
\label{eq:ideal word are means (simple)}
\vz{i} = \frac{1}{|\Zslice|} \sum_{z \in \Zslice} \Log_\mu(\uu_{z})
\end{equation}
Moreover, $\sum_{z_i \in \Z_i} \vz{i}=0$ for all $i=1, \dots, s$.
\end{prop}
This result tells us that each vector $\vz{i}$
in the optimal decomposition is the tangent mean of all the input compositions including the primitive $z_i$.
Moreover, the choice of the intrinsic mean as the point of tangency guarantees the uniqueness constraint is satisfied (see \cref{sec:proofs}).

\subsection{Decomposable Embeddings of Visual Inputs}
\label{sec:GDE from visual inputs}
Our framework holds for arbitrary manifolds and for any embedding map, hence being independent of the input modality. 
However, collections of natural visual data contain \emph{noise} and are \emph{sparse}. We account for these properties in our framework as presented in the following.

\subsubsection{Removing \emph{noise} from finite image sets}
We refer to \emph{noise} as information carried by images in addition to the composite concept of interest.
For example, an image from the tuple $z = \text{(red, car)}$ likely contains non-negligible extra information, e.g. a driver, a road, or a blue sky in the background.
This stems from the inherent ambiguity and non-uniqueness of visual signals. Most importantly, it is absent in text, for which it is easier to manually craft the string ``a red car'' ensuring no extra information.

\noindent\textbf{Problem formulation.} Since images contain noise in addition to represented concepts, we consider an input set $\encoder(\Z\times\Zextra)=\{\uu_{(z,e)}\suchthat (z,e)\in\Z\times\Zextra\}$ where each $z$ %
is represented by $k=|\Zextra|$ different image embeddings varying along the unknown noise dimension $\Zextra$. 
Also, different images may contain different amounts of noise. For each fixed $z$, we model this aspect with a probability distribution $\{\scorez\}_{e \in \Zextra}$ describing how well the elements in $\{\uu_{(z, e)}\}_{e \in \Zextra}$ represent their label $z$. In this setting, we want the decomposable set $\{\ut_z\}_{z \in \Z}$ minimizing the objective
\begin{equation}
\label{eq:geodesic optimization problem}
    \sum_{(z, e) \in \Z\times\Zextra} \scorez d_\M(\uu_{(z, e)}, \ut_z)^2,
\end{equation}
where the importance given to the approximation error for each input embedding is weighted according to the noise distribution.
The next result generalizes Proposition \ref{prop:best decomposable approx (simple)}, which addresses the special case $k \trim{3}=\trim{3}1$, and provides an easy-to-compute approximate solution to the problem.
\begin{restatable}{prop}{propdecomposition}
\label{prop:best decomposable approx (weighted)}
Let $\scorez$, $(z, e) \in \Z \times \Zextra$, be non-negative scalars such that $\sum_{e \in \Zextra} \scorez = 1$ for each $z \in \Z$, and let $\phi(\Z\times\Zextra)=\{\uu_{(z, e)} \suchthat (z, e) \in \Z\times\Zextra\} \subset \M$ be a set of embeddings with weighted intrinsic mean $\mu$ w.r.t. the weights $w_{(z,e)}=\scorez / \sum_{(z, e)}\scorez$. The minimization problem:
\begin{equation}
\label{eq:linear optimization problem}
\begin{aligned}
    &\argmin_{\{\ut_z\}}  \sum_{(z, e) \in \Z\times\Zextra} \scorez ||\Log_\mu(\uu_{(z, e)}) - \Log_\mu(\ut_z)||^2, \\ &\qquad s.t. \; \{\ut_z\} \mbox{ is geodesically decomposable}
\end{aligned}
\end{equation}
is solved by $\ut_z = \Exp_\mu(\vz{1} + \dots + \vz{s})$, where
\begin{equation}
\label{eq:ideal word are means (weighted)}
    \vz{i} = \frac{1}{|\Zslice|} \sum_{z \in \Zslice} \vv_{z}
, \quad
    \vv_z = \sum_{e \in \Zextra} \scorez \Log_\mu(\uu_{(z, e)})
\end{equation}
Moreover, $\sum_{z_i \in \Z_i} \vz{i}=0$ for all $i=1, \dots, s$.
\end{restatable}
\cref{fig:diagram method} visualizes the decomposition procedure. 
Notice that, using the same notation of the proposition, the vectors $\vv_z$ can be seen as a denoised tangent representation of the tuples in $\Z$ and the solution $\{\ut_z\}_{z \in \Z}$ to the weighted optimization problem corresponds to the decomposable approximation given by Proposition \ref{prop:best decomposable approx (simple)} applied to the denoised embeddings $\{\uu_z:=\Exp_\mu(\vv_z)\}_{z \in \Z}$.
Indeed, these have intrinsic mean equal to the weighted intrinsic mean $\mu$.
\begin{restatable}{lemma}{lemmasamecenter}
    Using the notation of Proposition \ref{prop:best decomposable approx (weighted)}, the set $\{\uu_z:=\Exp_\mu(\vv_z)\}_{z \in \Z}$ has intrinsic mean $\mu$.
\end{restatable}

\subsubsection{Dealing with \emph{sparsity} in finite image sets}
The previously described setup assumes that every $z\in\Z$ is represented by $k>0$ images. 
This requirement can be too restrictive in practice, because some combinations of primitives may not occur in real image collections. 
For example, if $\Z=\{\mbox{red, blue}\}\times\{\mbox{car, apple}\}$, there will probably be no pictures of a (blue, apple).
We refer to the absence of composite concepts as \emph{sparsity}.
Once more, please note that sparsity is not an issue with text, since strings can be manually crafted for any $z \in \Z$.

\noindent\textbf{Problem Formulation.} In general, in a labeled image collection, only a subset $\T\subset \Z\times\Zextra$ is available, and only a subgroup $\Z'\subset\Z$ of composite concepts is represented by at least one element in $\T$.
In this scenario, we obtain a decomposable approximation of $\phi(\T)$ by approximating the vector means in \cref{eq:ideal word are means (weighted)} with the mean of the available elements. 
The only requirement is that every primitive $z_i\in\Z_i$ $(i=1, \dots, s)$ appears in at least one tuple of $\Z'$.
Precisely, we first compute the weighted intrinsic mean $\mu$ of $\phi(\T)$ with weights $w_{(z,e)}=\scorez/\sum_{(z,e)\in\T}\scorez$, and then consider $\ut_z = \Exp_\mu(\vz{1} + \dots + \vz{s})$, where:
\begin{equation}
\label{eq:ideal words are means (sparse)}
    \vz{i} = \frac{1}{|\Z'(z_i)|} \sum_{z \in \Z'(z_i)}\hspace{-.1cm} \vv_{z}
,\ \
    \vv_z =\hspace{-.2cm} \sum_{\substack{e \in \Zextra \, s.t.\\(z,e) \in\T}} \hspace{-.1cm} \scorez \Log_\mu(\uu_{(z, e)})
\end{equation}
Note that the obtained decomposable set contains vector representations of all the concepts in $\Z$, including the unseen elements of $\Z \setminus \Z'$. 
The formulation in \eqref{eq:ideal words are means (sparse)} deals with all aspects mentioned so far: the manifold $\M$, noise, and sparsity.
In the next section, we use it to evaluate the compositional structure of real visual embeddings.

\noindent\textbf{Noise distribution.}
\label{sec:noise distribution}
The described setup requires the noise scores $\scorez$.
Given a collection of visual inputs $\T$ representing each label $z \in \Z'$ with $k_{z}>0$ elements, a simple choice is using uniform scores $\scorez=1/k_z$.
Alternatively, we propose using the CLIP image-to-text distribution $\scorez = \Prob((z,e)|y(z))$, where $y(z)$ is a text prompt for label $z\in\Z'$. This is the softmax of the scaled similarities
\begin{equation}
\label{eq:noise scores as clip scores}
    \Prob((z,e)|y(z)) = \frac{\exp({\uu_{(z,e)}^\top\uu_{y(z)}/t})}{\sum_{e}\exp({\uu_{(z,e)}^\top\uu_{y(z)}/t})}
\end{equation}
The temperature parameter $t$ is learned during training, but it can be tweaked to smooth or sharpen the distribution.
 
\section{Experimental Validation}
\label{sec:experiments}
We carry out experiments to analyze the decomposable properties of visual embeddings of VLMs. When not specified differently, we use the pre-trained CLIP ViT-L/14~\cite{radford2021learning}. We also consider CLIP ResNet50~\cite{radford2021learning} and SigLIP~\cite{Zhai2023siglip}. All considered models are from the OpenCLIP repository~\cite{openCLIP}.
We use images with attribute-object labels to represent sets of composite concepts of the form ${\Z=\attrs\times\objs}$. %

In this setup, we first assess the decomposable nature of small sets of embeddings inspecting their geometric arrangement according to Proposition \ref{prop:best decomposable approx (weighted)} (\cref{sec:pca}).
Then, we leverage the structured nature of the decomposed embeddings and experiment on the tasks of compositional classification (\cref{sec:comp_class}) and group robustness (\cref{sec:debiasing}). Finally, we visualize the approximate decomposable embeddings using a diffusion model (StableDiffusion v2.1~\cite{rombach2022high}) with the unCLIP technique~\cite{ramesh2022hierarchical} (\cref{sec:generative}).

\noindent\textbf{Attribute-object decomposition.}
We usually deal with sparse collections of visual inputs $\T$ where only a subset $\Z'$ of labels present at least one image.
Thus, we compute the embedding decomposition according to \cref{eq:ideal words are means (sparse)}: the optimal vectors $\ut_{(a,o)}=\Exp_\mu(\vv_a+\vv_o)$ are the combinations of the attribute directions $\vv_a=\frac{1}{\Z'(a)}\sum_o \vv_{(a,o)}$ and the object directions $\vv_o=\frac{1}{\Z'(o)}\sum_a \vv_{(a,o)}$, where the denoised representations $\vv_{(a,o)}$, $(a,o)\in\Z'$, are the mean tangent vectors within pairs.
For compositional classification and group robustness, we use the CLIP image-to-text probabilities as the noise distribution discussed in \cref{sec:noise distribution}. We finetune the temperature parameter (see \cref{sec:noise distribution appendix} for details).
In the other experiments, we utilize uniform scores.

\noindent\textbf{Datasets.}
We represent composite concepts with images from the training sets of diverse compositional datasets. %
We test compositional classification on the typical benchmark datasets UT-Zappos \cite{zappos} and MIT-states \cite{mitstates} with the splits from  \cite{Purushwalkam2019ModularNetworksForCZSL}. %
UT-Zappos contains images of shoes centered on a white background all sharing the
same orientation. 
There are 12 object classes referring to the footwear type and 16 attribute categories referring to the material.
MIT-states is a collection of natural objects in different states. The dataset contains 115 attribute categories and 245 object categories, generating a large number of possible combinations.

We test group robustness on the Waterbirds and CelebA datasets with the splits in~\cite{waterbirds}. These contain objects with spuriously correlated attributes, making them suitable for debiasing tasks.
Waterbirds contains images of two bird species $\objs \shorteq\{\mbox{waterbird, landbird}\}$ on two  types of background $\attrs \shorteq \{\mbox{land, water}\}$. 
We use the version of CelebA from~\cite{waterbirds} that contains close-ups photos of celebrities labeled with hair-color $\objs\shorteq\{\mbox{blonde, dark}\}$ and gender $\attrs\shorteq\{\mbox{male, female}\}$.
The data distribution over the four different groups is highly unbalanced in the train sets of these two datasets, implying spurious correlations. %

\subsection{Visualizing Compositional Embeddings}
\label{sec:pca} 
We evaluated the decomposability of the embeddings from a geometric perspective. We visualize lower-dimensional PCA projections of the tangent vectors $\{\vv_{(a, o)}\}$, considering that the denoised representations $\uu_{(a,o)}:=\Exp_\mu(\vv_{(a,o)})$ are geodesically decomposable if and only if their tangent directions are the vertices of a geometric shape with parallel faces. For example, decomposable sets of size $|\Z|=2\times2$ and $|\Z|=2\times3$ correspond to a parallelogram and a triangular prism, respectively.

\begin{figure}[t]
    \centering
        \subfloat{
            \includegraphics[width=\columnwidth]{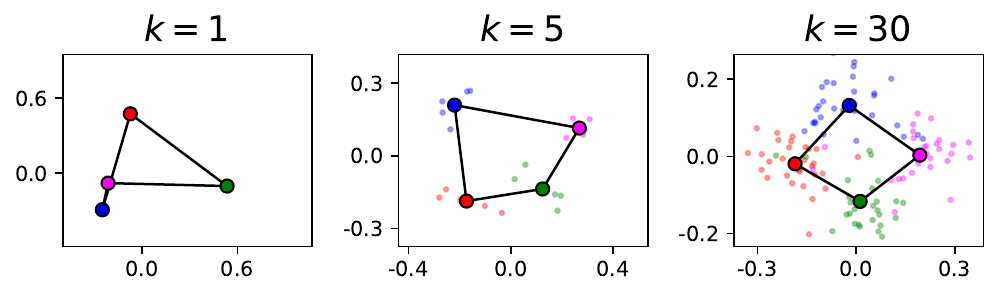}
        }

        \subfloat{
            \includegraphics[trim={0 10 0 0}, clip,width=\columnwidth]{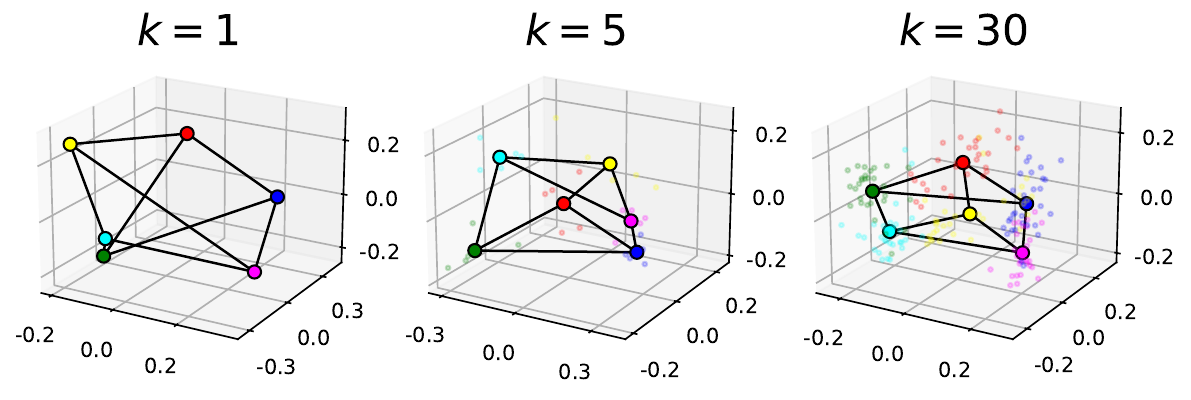}
            \label{fig:2x3 PGA projections}
        }
    \caption{(\textit{top}) 2-D projections of image embeddings representing the $2\times2$ composite labels of the Waterbirds dataset. (\textit{bottom}) 3-D projections of image embeddings representing $2\times3$ composite labels with images from the UT-Zappos dataset.
    Denoised pair representations (marked with a black contour) are computed with $k=1, 5, 30$ randomly selected images.}
    \label{fig:2x2 PGA projections}
\end{figure}

In~\cref{fig:2x2 PGA projections} we show the (first row) 2-D projection of image embeddings from the Waterbirds dataset, representing the four compositions of two attributes and two objects, and the (second row) 3-D projection of the two-by-three concepts in the set $\{\mbox{leather, suede}\}\times\{\mbox{boots ankle, boots knee high, shoes flats}\}$ of the UT-Zappos dataset. %
By increasing the number $k$ of images per pair, %
the noise is successfully removed and the resulting representations define shapes with parallel faces,  indicating approximate geodesical decomposability.
This highlights the importance of the denoising step and demonstrates compositional regularities of visual embeddings.

\subsection{Compositional Classification}
\label{sec:comp_class}
We perform compositional classification on the UT-Zappos and MIT-states datasets using the decomposable approximation of the train data as classifiers. 
This task serves to evaluate the generalization capabilities towards novel compositions of objects and states.
Specifically, we follow the standard generalized zero-shot evaluation protocol in both closed-world and open-world scenarios~\cite{mancini2021open}. 

We compute decomposable embeddings on a subset $\Z' \subset \Z$ of \emph{seen} pairs from the training set, while not all labels in the test set are in $\Z'$.
In the closed-world setting, the set of target labels $\Z^{test}\subset\Z$ contains only the pairs appearing in the dataset, while in the open-world framework, no prior knowledge is assumed and all the attribute-object combinations in $\Z^{test}=\Z$ are considered. 
Both settings require generalizing the prior knowledge about the primitives to understand the \textit{unseen} compositions in $\Z^{test}\setminus\Z'$. 
This operation is particularly challenging in the open-world scenario, where the more numerous novel compositions in the test set are a distraction for the predictor.

Our framework provides a straightforward solution to the complex problem of compositional classification. 
The geodesically decomposable set $\{\ut_{(a,o)}\}$ computed with the full train data $\T$ represents all the pairs, including the unseen ones. 
Thus we classify an image $x$ as $\argmax_{(a,o) \in \Z^{test}} \ut_{(a, o)}^\top \uu_x$.
We evaluate the prediction with the standard metrics~\cite{chao2016empirical, Purushwalkam2019ModularNetworksForCZSL}: attribute accuracy (ATTR), object accuracy (OBJ), best seen accuracy (SEEN), best unseen accuracy (UNSEEN), best harmonic mean (HM) between the seen and unseen accuracy and area under the seen-unseen curve (AUC).

\begin{table*}[!t]
    \centering
    \resizebox{\textwidth}{!}{
\begin{tabular}{cl ccccccc c ccccccc} 
\toprule 
 & & \multicolumn{7}{c}{\textsc{closed-world}} && \multicolumn{7}{c}{\textsc{open-world}}\\ 
\cline{3-9} \cline{11-17} 
\textbf{\textsc{Dataset}} & \textbf{\textsc{Method}} & \textsc{Attr} & \textsc{Obj} & \textsc{Seen} & \textsc{Unseen} & \textsc{HM} & \textsc{AUC} & $\rho$ && \textsc{A} & \textsc{O} & \textsc{S} & \textsc{U} & \textsc{H} & \textsc{AUC} & $\rho$\\ 
\midrule 
\cellcolor{white} & \textsc{CLIP} \cite{radford2021learning} & \itshape 24.1 & \itshape 58.3 & \itshape 11.9 & \itshape 45.7 & \itshape 15.3 & \itshape 4.4 & - && \itshape 18.8 & \itshape 57.4 & \itshape 11.9 & \itshape 23.8 & \itshape 12.0 & \itshape 2.3 & -\\ 
\cline{2-17}\cellcolor{white} & \textsc{LDE (text)}* \cite{trager2023linear} & 24.1 & 58.8 & 11.9 & 45.7 & 14.1 & 4.0 & 92.4 \% && \bfseries 19.2 & 57.2 & 11.9 & 20.0 & 11.1 & 1.9 & 83.2 \%\\ 
\rowcolor{myrowcolor} 
\cellcolor{white} & \textsc{GDE (text)} & \bfseries 25.3 & \bfseries 60.0 & \bfseries 17.0 & \bfseries 48.2 & \bfseries 18.9 & \bfseries 6.4 & \bfseries 146.6 \% && 18.7 & \bfseries 59.9 & \bfseries 17.0 & \bfseries 21.4 & \bfseries 12.2 & \bfseries 2.5 & \bfseries 111.1 \%\\ 
\cline{2-17}\cellcolor{white} & \textsc{LDE (image)} & 13.9 & 52.6 & 5.6 & 32.1 & 6.6 & 0.9 & 21.1 \% && 9.8 & 48.0 & 5.6 & 14.9 & 2.3 & 0.2 & 8.9 \%\\ 
\rowcolor{myrowcolor} 
\multirow{-5}{*}{\textsc{UT-Zappos}} \cellcolor{white} & \textsc{GDE (image)} & \bfseries 36.3 & \bfseries 64.1 & \bfseries 31.4 & \bfseries 55.9 & \bfseries 29.3 & \bfseries 13.9 & \bfseries 317.9 \% && \bfseries 28.6 & \bfseries 61.7 & \bfseries 31.3 & \bfseries 33.3 & \bfseries 19.0 & \bfseries 6.7 & \bfseries 293.5 \%\\ 
\midrule 
\cellcolor{white} & \textsc{CLIP} \cite{radford2021learning} & \itshape 33.0 & \itshape 52.1 & \itshape 30.6 & \itshape 45.3 & \itshape 26.3 & \itshape 11.1 & - && \itshape 15.6 & \itshape 47.7 & \itshape 30.6 & \itshape 8.3 & \itshape 8.4 & \itshape 1.7 & -\\ 
\cline{2-17}\cellcolor{white} & \textsc{LDE (text)}* \cite{trager2023linear} & 30.6 & 51.2 & 24.7 & 43.0 & 21.9 & 8.2 & 73.4 \% && 21.1 & \bfseries 50.7 & 24.7 & \bfseries 13.8 & 11.9 & 2.5 & 148.1 \%\\ 
\rowcolor{myrowcolor} 
\cellcolor{white} & \textsc{GDE (text)} & \bfseries 32.6 & \bfseries 51.7 & \bfseries 27.8 & \bfseries 45.2 & \bfseries 24.5 & \bfseries 10.0 & \bfseries 89.7 \% && \bfseries 21.3 & 49.9 & \bfseries 27.8 & 13.0 & \bfseries 12.1 & \bfseries 2.6 & \bfseries 158.5 \%\\ 
\cline{2-17}\cellcolor{white} & \textsc{LDE (image)} & 15.3 & 30.5 & 15.0 & 20.9 & 11.1 & 2.0 & 18.4 \% && 11.0 & 34.8 & 15.0 & 5.6 & 4.6 & 0.4 & 27.1 \%\\ 
\rowcolor{myrowcolor} 
\multirow{-5}{*}{\textsc{MIT-states}} \cellcolor{white} & \textsc{GDE (image)} & \bfseries 28.1 & \bfseries 45.3 & \bfseries 30.7 & \bfseries 36.1 & \bfseries 23.4 & \bfseries 8.6 & \bfseries 77.7 \% && \bfseries 18.5 & \bfseries 43.6 & \bfseries 29.7 & \bfseries 8.5 & \bfseries 9.3 & \bfseries 1.8 & \bfseries 106.6 \%\\ 
\bottomrule 
\end{tabular}
}\vspace{-2mm}
    \caption{Compositional classification results on the UT-Zappos and MIT-states datasets. Highest values within modality are in \textbf{bold} and ``*'' indicates that the results of our implementation are shown.}
    \label{tab:compositional classification CW, OW}
\end{table*}

\begin{table*}[!t]
    \centering
    \resizebox{\textwidth}{!}{
\begin{tabular}{cl ccccccc c ccccccc c ccccccc} 
\toprule 
 & & \multicolumn{7}{c}{\textsc{CLIP, RN50}} && \multicolumn{7}{c}{\textsc{CLIP, ViT-L/ 14}} && \multicolumn{7}{c}{\textsc{SigLIP, ViT-SO400M/ 14}}\\ 
\cline{3-9} \cline{11-17} \cline{19-25} 
\textbf{\textsc{Dataset}} & \textbf{\textsc{Method}} &\textsc{Attr} & \textsc{Obj} & \textsc{Seen} & \textsc{Unseen} & \textsc{HM} & \textsc{AUC} & $\rho$  && \textsc{Attr} & \textsc{Obj} & \textsc{Seen} & \textsc{Unseen} & \textsc{HM} & \textsc{AUC} & $\rho$ && \textsc{Attr} & \textsc{Obj} & \textsc{Seen} & \textsc{Unseen} & \textsc{HM} & \textsc{AUC} & $\rho$\\ 
\midrule 
\cellcolor{white} & \textsc{CLIP} \cite{radford2021learning} & \itshape 24.4 & \itshape 40.5 & \itshape 4.8 & \itshape 41.9 & \itshape 6.7 & \itshape 1.5 & - && \itshape 24.1 & \itshape 58.3 & \itshape 11.9 & \itshape 45.7 & \itshape 15.3 & \itshape 4.4 & - && \itshape 52.5 & \itshape 74.4 & \itshape 44.9 & \itshape 68.1 & \itshape 39.2 & \itshape 24.6 & -\\ 
\cline{2-25}\cellcolor{white} & \textsc{LDE (image)} & 15.1 & 44.6 & 3.2 & 21.7 & 4.7 & 0.5 & 33.5 \% && 13.9 & 52.6 & 5.6 & 32.1 & 6.6 & 0.9 & 21.1 \% && 21.4 & 50.3 & 7.0 & 42.1 & 8.2 & 1.6 & 6.5 \%\\ 
\rowcolor{myrowcolor} 
\multirow{-3}{*}{\textsc{UT-Zappos}} \cellcolor{white} & \textsc{GDE (image)} & \bfseries 28.2 & \bfseries 56.1 & \bfseries 24.0 & \bfseries 43.7 & \bfseries 23.7 & \bfseries 8.6 & \bfseries 578.5 \% && \bfseries 36.3 & \bfseries 64.1 & \bfseries 31.4 & \bfseries 55.9 & \bfseries 29.3 & \bfseries 13.9 & \bfseries 317.9 \% && \bfseries 48.1 & \bfseries 72.4 & \bfseries 42.5 & \bfseries 68.7 & \bfseries 41.3 & \bfseries 24.7 & \bfseries 100.4 \%\\ 
\midrule 
\cellcolor{white} & \textsc{CLIP} \cite{radford2021learning} & \itshape 26.6 & \itshape 42.3 & \itshape 23.5 & \itshape 35.2 & \itshape 19.4 & \itshape 6.2 & - && \itshape 33.0 & \itshape 52.1 & \itshape 30.6 & \itshape 45.3 & \itshape 26.3 & \itshape 11.1 & - && \itshape 45.9 & \itshape 61.2 & \itshape 43.8 & \itshape 58.1 & \itshape 39.7 & \itshape 22.2 & -\\ 
\cline{2-25}\cellcolor{white} & \textsc{LDE (image)} & 13.7 & 25.1 & 10.6 & 16.2 & 8.0 & 1.1 & 17.8 \% && 15.3 & 30.5 & 15.0 & 20.9 & 11.1 & 2.0 & 18.4 \% && 18.7 & 34.6 & 18.6 & 27.2 & 14.6 & 3.6 & 16.1 \%\\ 
\rowcolor{myrowcolor} 
\multirow{-3}{*}{\textsc{MIT-states}} \cellcolor{white} & \textsc{GDE (image)} & \bfseries 20.8 & \bfseries 34.4 & \bfseries 18.9 & \bfseries 25.1 & \bfseries 14.2 & \bfseries 3.4 & \bfseries 54.9 \% && \bfseries 28.1 & \bfseries 45.3 & \bfseries 30.7 & \bfseries 36.1 & \bfseries 23.4 & \bfseries 8.6 & \bfseries 77.7 \% && \bfseries 32.3 & \bfseries 50.3 & \bfseries 36.8 & \bfseries 40.8 & \bfseries 27.3 & \bfseries 11.9 & \bfseries 53.6 \%\\ 
\bottomrule 
\end{tabular}
}\vspace{-2mm}
    \caption{Ablation on backbone architecture in compositional classification, closed-world scenario.}
    \label{tab:ablation backbone CW}
\end{table*}

\noindent\textbf{Baselines.}
Our primary goal is to examine if the Geodesically Decomposable Embeddings (GDE) approximating the train data contain semantically meaningful information about the composite concepts they represent. 
We evaluate the relative performance $\rho$ (AUC ratio) obtained with decomposed embeddings w.r.t. the results achieved with the standard zero-shot baseline (CLIP) using the full-state embeddings 
(attribute-object labels ${(a, o) \in \Z^{test}}$ are represented by the text embedding of $\mbox{``An image of a } \{a\} \, \{o\}\mbox{''}$).

We investigate the importance of complying with data geometry and compare with the Linearly Decomposable Embeddings (LDE) proposed in~\cite{trager2023linear}, which we compute by setting $\M=\Real^n$ in our method, for both text and image modalities. We indicate the modality by adding ``\textsc{(text)}'' or ``\textsc{(image)}'' next to method names.
Decomposed text-embeddings are given by Proposition \ref{prop:best decomposable approx (simple)}, as noise and sparsity belong only to visual data.

\noindent\textbf{Results.}
Table \ref{tab:compositional classification CW, OW} reports the results in the closed-word and open-world settings.
In general, GDEs of visual data perform closely to the zero-shot full-state baseline, demonstrating they encode semantically meaningful information about the labels.
Interestingly, on the UT-Zappos dataset, they improve the standard zero-shot approach by a large margin. 
We attribute this gap to the fact that in UT-Zappos numerous representations are used for the computation of each primitive direction on average ($\sim$1400 per attribute, $\sim$1900 per object).
In contrast, the MIT-states dataset contains noisy annotations \cite{atzmon2020causal} and on average fewer representations to compute the primitives ($\sim$260 per attribute, $\sim$120 per object).
The decomposition shows robustness to sparsity, as indicated by the good open-world unseen accuracy on the MIT-states datasets, for which seen pairs are less than 5\% of the total.
LDE for visual data performs much worse than GDE on both datasets and when ablating the VLM backbone (see~\cref{tab:ablation backbone CW}).
This indicates that image embeddings are not closely linearly decomposable, and highlights the importance of respecting the data geometry when dealing with the extra complexity given by noise and sparsity. 
This verifies also on other geometries (see \cref{sec:hyperbolic results}).

\subsection{Group Robustness} 
\label{sec:debiasing}
\newcommand{\GG}{\mathcal{G}}
Pre-trained VLMs %
produce biased representations, leading to zero-shot classifiers not robust to group shifts~\cite{zhang2022contrastive}. 
Our framework offers a training-free method to compute unbiased embeddings. 
We evaluate it on the group robustness benchmark presented in \cite{waterbirds}, which requires classifying an image without leveraging spurious correlations. 
In this setting, a set of target classes $\objs$ has spurious correlations with a set of attributes $\attrs$ due to the highly unbalanced data distribution over the \emph{groups} in $\GG = \attrs \times \objs$. The goal is to obtain an object classifier that does not exploit spurious correlations, improving the average accuracy over all the groups (AVG) while keeping the (GAP) on the worst group accuracy (WG) small.
We use the object embeddings ${\ut_o:=\Exp_\mu(\vv_o)}$, ${o \in \Z_{obj}} $ computed with our method to evaluate the group robustness performance. Intuitively, these embed only object representations that are not correlated with attribute-related spurious features.
We thus predict the object class of an image $x$ as $\argmax_{o\in\Z_{obj}}\ut_o^\top \uu_x$.

\noindent\textbf{Baselines.} 
In addition to the zero-shot CLIP and {\LDE} method, we include two standard baselines that use labeled data, namely Empirical Risk Minimization (ERM) with linear probing \cite{kumar2022fine} and ERM with feature adapters \cite{gao2024clip}. Furthermore, we compare with two recent methods improving the performance of VLMs, Deep Feature Reweighting (DFR) \cite{kirichenko2023last} and Contrastive Adapters (CA) \cite{zhang2022contrastive}, and with FairerCLIP \cite{dehdashtian2024fairerclip} that performs debiasing of the frozen CLIP representations in the training-free setting like our method.

\begin{figure}[!t]\vspace{-2mm}
    \centering
    \includegraphics[trim={0 7 0 0}, clip, width=0.95\linewidth]{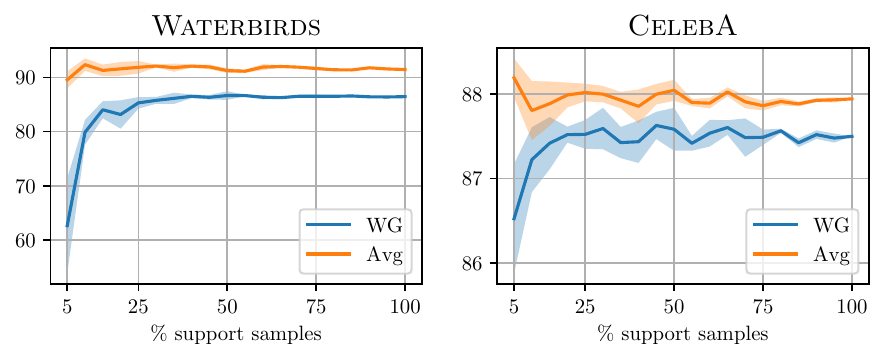}
    \vspace{-2mm}
    \caption{Data efficiency of \GDE on the group robustness benchmark. Subsets of the full support set are sampled keeping the group ratios fixed. The shaded confidence band shows the standard deviation over five experiments.}
    \label{fig:ablate k debiasing}
\end{figure}

\noindent\textbf{Results.}
Table \ref{tab:debiasing} reports the results on the group robustness benchmarks. 
{\ours} considerably outperforms CLIP and {\LDE}, with an increase of WG accuracy on the Waterbirds and CelebA datasets of about 42 and 21.8, respectively. 
This indicates that our method effectively decomposes the embeddings of object and attribute primitives, producing robust classifiers.
Notably, it achieves state-of-the-art WG accuracy and smaller Gap compared to all other methods that use labeled data, including the task-specific FairerCLIP.  \ours is thus an effective training-free solution to compute unbiased embeddings.
Furthermore, {\ours} demonstrates remarkable data-efficiency performance, achieving high results using limited amount of data (see~\cref{fig:ablate k debiasing}). For example, when using 25\% of the full train samples (randomly selected keeping group ratios fixed) the WG decreases less than 1\% on both the Waterbirds and CelebA datasets.

\begin{table}[]
    \centering
    \setlength{\tabcolsep}{4pt}
    \resizebox{\columnwidth}{!}{
    \begin{tabular}{l ccc c ccc} 
       \toprule 
       & \multicolumn{3}{c}{\textsc{Waterbirds}} && \multicolumn{3}{c}{\textsc{CelebA}} \\ 
       \cline{2-4} \cline{6-8} 
       \textbf{\textsc{Method}} & \textsc{WG} & \textsc{Avg} & \textsc{Gap} && \textsc{WG} & \textsc{Avg} & \textsc{Gap} \\ 
       \midrule 
       \textsc{CLIP} \cite{radford2021learning} & 44.4 & 84.3 & 40.0 && 74.4 & 86.9 & 12.4 \\
       \textsc{{\LDE (text)}}  \cite{trager2023linear}
       & 64.6 & 88.0 & 23.3 && 83.9 & 85.5 & 1.6\\
       \hline
       \textsc{ERM Linear Probe} \cite{kumar2022fine}
       & 65.4 & 97.7 & 32.3 && 30.4 & \bfseries 94.6 & 64.2 \\
       \textsc{ERM Adapter} \cite{gao2024clip}
       & 76.1 & \bfseries 97.8 & 21.7 && 40.0 & 94.3 & 54.3 \\
       \textsc{DFR (Subsample)} \cite{kirichenko2023last}
       & 58.8 & 95.9 & 37.1 && 78.7 & 91.8 & 13.1 \\
       \textsc{DFR (Upsample)} \cite{kirichenko2023last}
       & 66.5 & 96.4 & 29.8 && 83.9 & 91.2 & 7.2  \\
       \textsc{CA} \cite{zhang2022contrastive}
       & 85.3 & 94.5 & 9.3  && 83.9 & 90.4 & 6.4  \\
        \textsc{FairerCLIP} \cite{dehdashtian2024fairerclip}
       & 86.0 & 92.2 & 6.1 && 85.2 & 87.8 & 2.5 \\
       \rowcolor{myrowcolor}
        \textsc{\GDE (image)} & \bfseries 86.4 & 91.5 & \bfseries 5.0 && \bfseries 87.5 & 87.9 & \bfseries 0.4 \\
       \bottomrule
    \end{tabular}
    }\vspace{-2mm}
    \caption{Comparison of results on group robustness.} %
    \label{tab:debiasing}
\end{table}

\subsection{Visualize Decomposable Approximations}
\label{sec:generative} 
We visualize the decomposed visual embeddings %
using a diffusion model implementing the unCLIP mechanism (StableDiffusion v2.1)~\cite{rombach2022high, ramesh2022hierarchical}, trained to invert the CLIP image encoder by conditioning the generative process with the image embeddings. %
We invert the decomposable vectors $\ut_{(a,o)}$ obtained in previous experiments. In this way, we can qualitatively examine the information they contain.

\begin{figure}[!t]
    \centering
    \subfloat{
        \includegraphics[trim={0 0 0 7}, clip,width=\columnwidth]{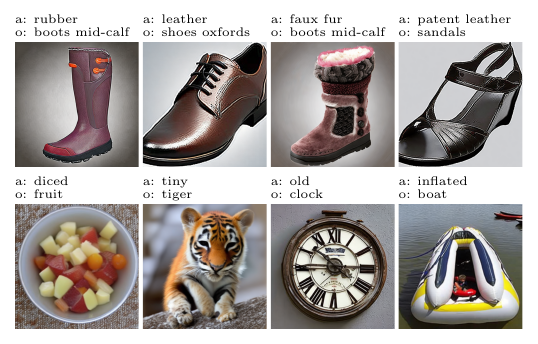}
        \label{fig:SD attr-obj pairs}
    }

    \vspace{-.3cm}
    
    \subfloat{
        \includegraphics[trim={0 0 0 2}, clip, width=\columnwidth]{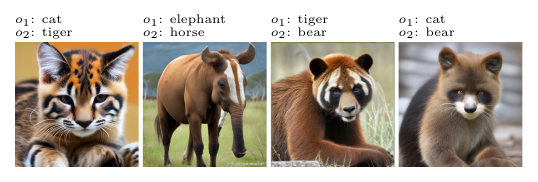}
        \label{fig:SD animals}
    }
    \vspace{-3mm}
    \caption{Attribute-object pairs generated using decomposed embeddings with StableDiffusion for the UT-Zappos (\textit{first row}) and MIT-states (\textit{second row}) datasets.  The two leftmost labels are seen pairs, while the two right-most are unseen pairs. We also generate object-object pairs (\textit{third row}) blending animal species.}
    \label{fig:SDgenerated}
\end{figure}

In~\cref{fig:SDgenerated} we show some generated images for object-attribute pairs where the attribute is not the most common state of the object (i.e. we avoid common pairs like ``green broccoli'' or ``big elephant''), observing whether the generated image correctly represents the full label and not just the attribute/object. %
The generated images well represent both the object and the attribute of the label, with no difference in the quality of the outputs from seen (two leftmost columns) and unseen pairs (two rightmost columns).
This emphasizes the generalization properties of our decomposable image embeddings, with potential to be applied in practical tasks like augmenting compositional sparse datasets. %

The modularity of the decomposable structures allows representing the composition of two objects $o_1, o_2 \in \objs$ as %
$\Exp_\mu(\vv_{o_1}+\vv_{o_2})$. 
Inspired by~\cite{longari2024blend}, we experiment with blending different animal species (\cref{fig:SDgenerated}, third row). The generated images portray photorealistic creatures with features of the two input species. This further highlights the power and versatility of the proposed framework. More generated images are in \cref{sec:extra generated images}.

\section{Conclusion}
We investigated the emergence of compositional structures within the image latent space of vision-language models and demonstrated that visual embeddings also exhibit a degree of compositionality similar to that of textual representations. 
We proposed a training-free framework, Geodesically Decomposable Embeddings (GDE), designed to address the noisy and sparse nature of image data. 
GDE decomposes visual representations as a geometry-aware combination of optimal directions representing primitive concepts.
We demonstrated that these composed representations encode complex concepts and are effective in several tasks, including compositional classification and group robustness. Notably, GDE presents more robust abilities to perform compositionality than existing approaches based on linear decomposition of latent spaces, contributing to higher results in group robustness than existing task-specific methods.
We believe this work contributes to achieving better interpretability and controllability of modern VLMs.

{\small
\vspace{2pt}\noindent\textbf{Acknowledgements.}
This work was sponsored by the ERJU project, the EU Horizon project ELIAS (No. 101120237), Ministero delle Imprese e del Made in Italy (IPCEI Cloud DM 27 giugno 2022 – IPCEI-CL-0000007), and the FAIR - Future AI Research (PE00000013), funded by NextGeneration EU.
The authors acknowledge the CINECA award under the ISCRA initiative for the availability of high-performance computing resources and support.
}
{
    \small
    \bibliographystyle{ieeenat_fullname}
    \bibliography{main}
}

\appendix

\maketitlesupplementary

\newcommand{\vbZ}[1]{\Bar{\vv}_{\Z_{#1}}}
\newcommand{\vt}{\Tilde{\vv}}
\newcommand{\Id}{I_d}
\newcommand{\Ball}{\mathcal{B}}

\noindent
In this Supplementary Material, we provide additional details on Riemannian manifolds in \cref{sec:Riemannian manifold details}, we prove the theoretical results our framework builds upon in \cref{sec:proofs}, we describe extra information of our implementation in \cref{sec:experimental details} and we present further experimental results in \cref{sec:additional results}.

\section{Details on Riemannian Manifold}\label{sec:Riemannian manifold details}
We discuss some details of the tools we used in our framework to deal with the geometry of a data manifold $\M$. 
In the following, we focus on the spherical case $\M=\Sphere^{d-1}$, which applies to the case with normalized embeddings.

\subsection{Closed form solutions}
The exponential and logarithmic maps can be expressed in closed form on the unit sphere $\Sphere^{d-1}$. For any point of tangency $\p \in \Sphere^{d-1}$, we have
\begin{equation}
\label{eq:exp map sphere}
    \Exp_\p(\vvv) = \cos(||\vvv||)\p + \sin(||\vvv||) \frac{\vvv}{||\vvv||}, \quad \vvv \in T_\p\Sphere^{d-1}
\end{equation}
and
\begin{equation}
\label{eq:log map sphere}
    \Log_\p(\q) = \theta \frac{(\Id - \p \p^\top)(\q - \p)}{||(\Id - \p \p^\top)(\q - \p)||}, \quad \q \in \Sphere^{d-1}
\end{equation}
where $\theta \short{=}{0} \arccos(\q^\top \p)$ and $\Id \short{\in}{0} \Real^{d \times d}$ is the identity matrix.\footnote{
To be precise, the logarithmic map is defined on $\M\setminus \mathcal{C}_\mu$, where $\mathcal{C}_\mu$ is called the \emph{cut locus} of $\mu$. We do not stress this detail because it is well known that $\mathcal{C}_\mu$ has measure zero on $\M$%
.
On the unit sphere $\Sphere^{d-1}$, the cut locus of any point $\mu$ is its antipode $-\mu$.
}

\subsection{Intrinsic Mean}\label{sec:intrinsic mean details}
\paragraph{Existence, uniqueness, and characterization.}
The (weighted) intrinsic mean $\mu$ of a set of points $\{\uu_i\}_{i=1}^N$, which is defined as the solution of a minimization problem, %
is not necessarily unique.
For example, on $\Sphere^2$ all the points on the equator minimize the average distance from the north and south poles.
But, existence and uniqueness are guaranteed if the points live inside the same geodesic ball $\Ball_o(r):=\{\uu\in\M\suchthat d_\M(o, \uu)<r\}$ of radius $r>0$ small enough~\cite{afsari2011riemannian}.
Under the same condition, we also have that $\mu$ is the unique point on $\M$ centering the logarithmic map of the input points, \ie such that $\sum_{i=1}^N w_i \, \Log_\mu(\uu_i) = 0$.
We will refer to this property as the \emph{characterization of the intrinsic mean}.
For the unit sphere $\Sphere^{d-1}$, the closeness assumption is satisfied for any $r<\pi/2$. 
Note that we can expect this condition to be verified by the normalized embeddings of a neural encoder because of the cone effect~\cite{liang2022mind}.

\paragraph{Computation by gradient descent.}
\begin{algorithm}[t]
\caption{Intrinsic mean}\label{alg: intrinsic mean}
\textbf{Input:} $\uu_1, \dots, \uu_N \in \M, w_1, \dots, w_N \in \Delta_N, \mu_0 \in \M$ \\
\textbf{Output:} the intrinsic mean $\mu \in \M$
\begin{algorithmic}
\Repeat
\State $\delta_{\mu} = \eta \sum_{i=1}^N w_i \, \Log_{\mu_j}(\uu_i)$
    \State $\mu_{j+1} = \Exp_{\mu_j}(\delta_{\mu})$
\Until{$||\delta_{\mu}||<\epsilon$}
\end{algorithmic}
\end{algorithm}
Computing the intrinsic mean $\mu$ of a weighted set of points requires minimizing the objective function
\begin{equation}\label{eq: objective intr mean}
    f(\uu) = \frac{1}{2} \sum_{i=1}^N w_i \, d_\M(\uu, \uu_i)^2, \quad \uu\in\M
\end{equation}
This can be done with a gradient descent algorithm \cite{pennec1999probabilities}. 
Indeed, it can be shown that  (\ref{eq: objective intr mean}) has gradient 
\begin{equation}\label{eq: gradient intrinsic mean objective}
    \nabla f(\uu)  = - \sum_{i=1}^N w_i \, \Log_\uu(\uu_i),\quad \uu\in\M
\end{equation}
\cref{alg: intrinsic mean} shows the pseudocode for the gradient descent procedure. 
At each iteration, the new approximation $\mu_{j+1}$ is obtained by first moving in the opposite direction of the gradient and then mapping on the manifold with the exponential map centered in $\mu_j$. The cycle stops when the norm of the update is smaller than a fixed small value $\epsilon\short{>}{1}0$.
Usually, the starting value $\mu_0 \in \M$ is chosen among the input points, which live on the manifold. Otherwise, in the special case $\M = \Sphere^{d-1}$, a good choice is the normalized (weighted) arithmetic mean 
$
    \mu_0 = \sum_{i=1}^N w_i \,\uu_i / 
    \Vert \sum_{i=1}^N w_i \, \uu_i \Vert
$.
The learning rate $\eta$ has to be carefully chosen to guarantee convergence. It has been shown that setting $\eta=1$ is sufficient for spheres \cite{afsari2013convergence}.

\section{Proofs}
\label{sec:proofs}
We provide the proof of the theoretical results stated in the methodology chapter.
We omit the proof of Proposition 1 %
because it is the same  of the more general Proposition \ref{prop:best decomposable approx (weighted)} in the special case when $|\Zextra|=1$. In the following, we assume that a given composite concept $z\in \Z$ is the tuple $z=(z_1, \dots, z_s)$.

\lemmauniqueness*

\begin{proof}
Let $\phi(\Z)\shorteq\{\uu_z\}$ be a geodesically decomposable set with tangent projections $\vv_z \shorteq \Log_\mu(\uu_z)$ decomposed as $\vv_z \shorteq \vz{1}' + \dots + \vz{s}'$.
Indicating $\vbZ{i}=\frac{1}{|\Z_i|}\sum_{z_i \in \Z_i} \vz{i}'$, we now show that the searched directions are $\vz{i}=\vz{i}'-\vbZ{i}$ $(i=1, \dots, s)$. 
The centering constrain $\sum_{z_i \in \Z_i} \vv_{z_i}=0$ immediately follows from the definition. 
Then, we observe that $\sum_{i} \vbZ{i} =\frac{1}{|\Z|}\sum_{z \in \Z} \vv_{z}=0$ for the characterization of the intrinsic mean. This implies \cref{eq:uz decomposition} is satisfied:
\begin{equation}
\begin{aligned}
    \vv_z &= \vz{1}' + \dots + \vz{s}' \\
    &= (\vbZ{1} + \dots + \vbZ{s}) + \vz{1} + \dots + \vz{s} \\
    &=\vz{1} + \dots + \vz{s}\\
\end{aligned}
\end{equation} 
To show uniqueness, we demonstrate the $\vz{i}$ are uniquely determined by the original vectors $\vv_z$:

\begin{equation}
\begin{aligned}
    \vz{i} &= \vz{i}'-\vbZ{i} \\
        &= \vz{i}' + \sum_{j \ne i} \vbZ{j} \\
        &= \frac{1}{|\Zslice|} \sum_{z \in \Zslice} (\vz{1}' + \dots + \vz{s}') \\
        &= \frac{1}{|\Zslice|} \sum_{z \in \Zslice} \vv_z
\end{aligned}
\end{equation}
\end{proof}

\propdecomposition*

\begin{proof}
We start by observing that, if $\{\ut_z\}$ is a geodesically decomposable set with intrinsic mean $\mu'$, then, following the proof of Lemma \ref{lemma:decomposition is unique}, we can write its tangent projection $\vt_z=\Log_\mu(\ut_z) \in T_\mu\M$ as $\vt_z = \vv_0 + \vz{1} + \dots + \vz{s}$ where $\sum_{z_i\in\Z_i}\vz{i}=0$ and $\vv_0=\sum_i \vbZ{i}=\frac{1}{|\Z|} \sum_{z}\vt_z$. 
Note that $\mu'=\mu$ if and only if $\vv_0=0$.
Now, in the setting of the statement, we indicate $\vv_{(z,e)}\shorteq\Log_\mu(\uu_{(z,e)})$ and rephrase the objective in \cref{eq:linear optimization problem} as finding the vectors $\vv_0$, $\vz{i} \in T_\mu \M$, $z_i \in \Z_i$ $(i=1, \dots,s)$ minimizing
\begin{equation}
\label{eq:proof objective}
    \frac{1}{2}\sum_{\substack{(z, e) \\ \in \Z\times\Zextra}} \scorez ||\vv_{(z, e)} - (\vv_0 + \vz{1} + \dots + \vz{s})||^2
\end{equation}
Observing that $\sum_{z\in\Z}\vz{i}\shorteq\sum_i\frac{|\Z|}{|\Z_i|}\sum_{z_i\in\Z_i}\vz{i}\shorteq0$, the derivative of (\ref{eq:proof objective}) with respect to $\vv_0$ is
\begin{equation}
\begin{aligned}
     &\sum_{z \in \Z} \sum_{e \in \Zextra} \scorez (\vv_{(z, e)} - (\vv_0+\vz{1} + \dots + \vz{s})) \\
     &=\sum_{z \in \Z} (\vv_z - \vv_0), \\
\end{aligned}
\end{equation}
where $\vv_z=\sum_{e \in \Zextra} \scorez \vv_{(z, e)}$.
Setting this equal to zero gives $\vv_0 = \frac{1}{|\Z|} \sum_z \vv_{z}=\sum_{(z,e)} w_{(z,e)} \vv_{(z,e)}=0$. 
Here the last equality follows from the characterization of the intrinsic mean and it implies the intrinsic mean of the solution is $\mu$.
The derivative with respect to a fixed $\vz{i}$ is:
\begin{equation}
\begin{aligned}
     &\sum_{z \in \Zslice} \sum_{e \in \Zextra} \scorez (\vv_{(z, e)} - (\vz{1} + \dots + \vz{s})) \\
     &=\sum_{z \in \Zslice} (\vv_z - \vz{i}) \\
\end{aligned}
\end{equation}
Setting this equal to zero gives $\vz{i}=\frac{1}{|\Zslice|}\sum_{z\in\Zslice} \vv_z$.
\end{proof}

\lemmasamecenter*

\begin{proof}
As observed in the proof of Proposition \ref{prop:best decomposable approx (weighted)}, we have $\frac{1}{|\Z|}\sum_{z} \vv_z = 0$, implying the weighted mean $\mu$ is the intrinsic mean of $\{\uu_z\}_{z \in \Z}$.
\end{proof}

\section{Experimental Details}
\label{sec:experimental details}
\subsection{Closeness assumption}
We numerically verify the closeness assumption, discussed in \cref{sec:intrinsic mean details}, which guarantees the existence and uniqueness of the intrinsic mean.
Given a set of points on $\Sphere^{d-1}$, a good guess for the center $o\in\M$ of a small geodesic ball $\Ball_o(r)$ containing them is their normalized arithmetic mean $\mu_0$. 
So, for all the sets of embeddings used in our experiments, we verify their maximum intrinsic distance (\ie angle) from $\mu_0$ is smaller than $\pi / 2$. 
In \cref{tab:closeness assumption} we show some statistics of the distances computed with the embeddings from the default model CLIP ViT-L\textbackslash 14 used in the experiments.

\begin{table}[]
    \centering
\resizebox{\linewidth}{!}{
\begin{tabular}{c ccc c ccc} 
   \toprule 
   & \multicolumn{3}{c}{Image Embeddings} && \multicolumn{3}{c}{Text Embeddings} \\ 
   \cline{2-4} \cline{6-8} 
   & avg & max & $r<\pi/2$ && avg & max & $r<\pi / 2$ \\ 
   \hline 
   \textsc{UT-Zappos}  & 0.49 & 1.0  & \checkmark && 0.56 & 0.75 & \checkmark \\ 
   \textsc{MIT-states} & 0.78 & 1.4  & \checkmark && 0.67 & 1.14 & \checkmark \\ 
   \textsc{Waterbirds} & 0.63 & 1.03 & \checkmark && 0.41 & 0.48 & \checkmark \\ 
   \textsc{CelebA}     & 0.75 & 1.15 & \checkmark && 0.4  & 0.43 & \checkmark \\ 
   \bottomrule
\end{tabular}
}
\vspace{-2pt}

    \caption{Statististics of distances from embeddings to their normalized arithmetic mean. The closeness assumption is verified if all the embeddings are within a radius $r<\pi / 2\approx 1.57$.}
    \label{tab:closeness assumption}
\end{table}

\subsection{Noise distribution}\label{sec:noise distribution appendix}
\paragraph{Temperature selection.}
When performing compositional classification and group robustness, we use the image-to-text distribution $\Prob((z,e)|y(z))$ defined by the VLM as the noise distribution.
For CLIP, this is given by the softmax activations described in the main paper
and it depends on the temperature parameter $t \in (0,+\infty)$.
For each dataset, we select the value for $t$ by performing a grid search on the validation set. We optimize the AUC metric for compositional classification and the WG accuracy for group robustness.

\paragraph{SigLIP sigmoid probabilities.}
Differently from the original CLIP, SigLIP uses a sigmoid-based loss processing every image-text pair independently and it defines the pair-specific probabilities
\begin{equation}
\label{eq:SigLIP sigmoid probs}
\Prob((z,e)|y(z))=\frac{1}{1+\exp(-\uu_{(z,e)}^\top\uu_{y(z)}/t - b)}
\end{equation}
When considering SigLIP embeddings, we use the noise distribution $\scorez \propto \Prob((z,e)|y(z))$ proportional to the pair-specific sigmoid probabilities.
We select the temperature parameter $t$ as described for the CLIP model while keeping the \emph{logit bias} $b\in\Real$ equal to the learned value ($b\approx-16.5$).

\subsection{Text prompts}
For the UT-Zappos and MIT-states datasets, we consider the same text prompts used in~\cite{trager2023linear}. 
Attribute-object pair $(a, o)$ is described by $y(a,o)=\mbox{``an image of a } \{a\} \, \{o\}\mbox{''}$ where $\{a\}$, $\{o\}$ are the lower-case original category names. For UT-Zappos, every dot character is substituted with a space (``Synthetic Boots.Ankle'' $\to$ ``synthetic boots ankle'').
We use these prompts both when decomposing text embeddings and when computing the image-to-text probabilities defining the noise distribution.

For the Waterbirds and CelebA datasets, we consider the text prompts defined in~\cite{trager2023linear, chuang2023debiasing}. 
These are obtained by representing each spurious attribute and each target class with the captions in \cref{tab:waterbirds prompts,tab:celebA prompts}. Then, prepending the spurious prompts to the class prompts produces $k=4$ and $k=3$ textual descriptions for each composite group in the Waterbirds and CelebA datasets, respectively.
We compute the image-to-text probabilities for the noise distribution using the decomposable text embeddings $\ut_{y(z)}$, $z \in \Z$, given by Proposition \ref{prop:best decomposable approx (weighted)} applied to the input embeddings. 
Note indeed that they can be written as $\{\uu_{y(z,e)}\suchthat (z,e) \in \Z\times \Zextra\}$, where $\Zextra$ is a ``prompt template'' dimension.

\begin{table}[]
    \centering
\resizebox{\linewidth}{!}{
\begin{tabular}{c}
\toprule
\textbf{Class prompt}  \\
\hline
This is a picture of a landbird. \\
This is a picture of a waterbird. \\
\midrule
\textbf{Spurious attribute prompt} \\
\hline
This is a land background.     \qquad This is a water background.   \\
This is a picture of a forest. \qquad This is a picture of a beach. \\
This is a picture of a moutain.\qquad This is a picture of an ocean.\\
This is a picture of a wood.   \qquad This is a picture of a port.  \\
\bottomrule
\end{tabular}
}
    \caption{The text prompts from \cite{chuang2023debiasing} for the Waterbirds dataset.}
    \label{tab:waterbirds prompts}
\end{table}

\begin{table}[]
    \centering
\resizebox{\linewidth}{!}{
\begin{tabular}{c}
\toprule
\textbf{Class prompt}  \\
\hline
A photo of a celebrity with dark hair.  \\
A photo of a celebrity with blond hair. \\
\midrule
\textbf{Spurious attribute prompt} \\
\hline
A photo of a male.           \qquad A photo of a female.          \\
A photo of a male celebrity. \qquad A photo of a female celebrity.\\
A photo of a man.            \qquad  A photo of a woman.          \\
\bottomrule
\end{tabular}
}
    \caption{The text prompts from \cite{chuang2023debiasing} for the CelebA dataset.}
    \label{tab:celebA prompts}
\end{table}

\section{Additional Results}
\label{sec:additional results}

\subsection{Ablation: noise distribution}
Our decomposition method (GDE) computes the noise distribution using CLIP scores with a custom temperature parameter.
In \cref{tab:ablate CLIP scores CW}, we compare GDE against the decomposition obtained when using a uniform noise distribution ($\mbox{GDE}_u$) in the task of compositional classification.
While the simpler $\mbox{GDE}_u$ performs well compared to the zero-shot baseline, leveraging the non-uniform noise distribution from the CLIP scores always improves performance.

\begin{table}[]
    \centering
\resizebox{\linewidth}{!}{
\begin{tabular}{cl ccccccc} 
\toprule 
\textbf{\textsc{Dataset}} & \textbf{\textsc{Method}} &\textsc{Attr} & \textsc{Obj} & \textsc{Seen} & \textsc{Unseen} & \textsc{HM} & \textsc{AUC} & $\rho$\\ 
\midrule 
\cellcolor{white} & \textsc{CLIP} \cite{radford2021learning} & \itshape 24.1 & \itshape 58.3 & \itshape 11.9 & \itshape 45.7 & \itshape 15.3 & \itshape 4.4 & -\\ 
\cline{2-9} 
\cellcolor{white} & $\textsc{GDE}_u$ \textsc{(image)} & 36.2 & 63.8 & 30.9 & 55.6 & 29.0 & 13.6 & 310.8 \%\\ 
\rowcolor{myrowcolor} 
\multirow{-3}{*}{\textsc{UT-Zappos}} \cellcolor{white} & \textsc{GDE (image)} & \bfseries 36.3 & \bfseries 64.1 & \bfseries 31.4 & \bfseries 55.9 & \bfseries 29.3 & \bfseries 13.9 & \bfseries 317.9 \%\\ 
\midrule 
\cellcolor{white} & \textsc{CLIP} \cite{radford2021learning} & \itshape 33.0 & \itshape 52.1 & \itshape 30.6 & \itshape 45.3 & \itshape 26.3 & \itshape 11.1 & -\\ 
\cline{2-9} 
\cellcolor{white} & $\textsc{GDE}_u$ \textsc{(image)} & 27.7 & 44.3 & 30.4 & 35.0 & 22.9 & 8.2 & 74.3 \%\\ 
\rowcolor{myrowcolor} 
\multirow{-3}{*}{\textsc{MIT-states}} \cellcolor{white} & \textsc{GDE (image)} & \bfseries 28.1 & \bfseries 45.3 & \bfseries 30.7 & \bfseries 36.1 & \bfseries 23.4 & \bfseries 8.6 & \bfseries 77.7 \%\\ 
\bottomrule 
\end{tabular}
}

    \caption{Results of ablating the use of CLIP scores as the noise distribution in compositional classification, closed-world scenario.}
    \label{tab:ablate CLIP scores CW}
\end{table}

\subsection{ Decomposing hyperbolic representations}
\label{sec:hyperbolic results}
We investigate the compositional properties of visual representations on different geometries than the CLIP's hyper sphere. Specifically, we perform compositional classification of the pre-trained MERU ViT-L-16 \cite{desai2023hyperbolic} 
embeddings, which are points on the \emph{Lorentz model}:
\begin{equation}
\mathcal{L}^d=
\{\uu \in \Real^{d+1}|\langle\uu, \uu\rangle_\mathcal{L}=-1/c\}.
\end{equation}
Here $\langle\cdot,\cdot\rangle_\mathcal{L}$ is the Lorentzian inner product and the parameter $c>0$ is learned during pre-training. The exponential and logarithmic maps have a closed form solution for the hyperboloid $\mathcal{L}^d$ \cite{desai2023hyperbolic}, enabling a simple application of the GDE framework also in this setting.
Results in \cref{tab:classification MERU (CW)} show that, as observed for CLIP spherical embeddings, the GDEs of MERU's hyperbolic representations contain semantically meaningful information of the concepts they represent.
Moreover, the significantly lower performance of LDE highlights the importance of GDE's geometry awareness also in this non-spherical setup.

\begin{table}[t]
    \centering
\resizebox{\linewidth}{!}{
\begin{tabular}{cl ccccccc} 
\toprule 
\textbf{\textsc{Dataset}} & \textbf{\textsc{Method}} &\textsc{Attr} & \textsc{Obj} & \textsc{Seen} & \textsc{Unseen} & \textsc{HM} & \textsc{AUC} & $\rho$\\ 
\midrule 
\cellcolor{white} & \textsc{MERU} [\textcolor{cvprblue}{68}] & \itshape 17.4 & \itshape 26.7 & \itshape 11.1 & \itshape 16.0 & \itshape 9.6 & \itshape 1.4 & -\\ 
\cline{2-9} 
\cellcolor{white} & \textsc{LDE (image)} & 13.8 & 40.7 & 4.6 & 21.8 & 5.1 & 0.7 & 52.7 \%\\ 
\rowcolor{myrowcolor} 
\multirow{-3}{*}{\textsc{UT-Zappos}} \cellcolor{white} & \textsc{GDE (image)} & \bfseries 22.9 & \bfseries 49.7 & \bfseries 15.2 & \bfseries 40.3 & \bfseries 16.0 & \bfseries 4.7 & \bfseries 340.4 \%\\ 
\midrule 
\cellcolor{white} & \textsc{MERU} [\textcolor{cvprblue}{68}] & \itshape 17.7 & \itshape 34.4 & \itshape 15.8 & \itshape 27.0 & \itshape 13.2 & \itshape 3.1 & -\\ 
\cline{2-9} 
\cellcolor{white} & \textsc{LDE (image)} & 13.7 & 26.7 & 11.2 & 19.1 & 9.1 & 1.4 & 46.1 \%\\ 
\rowcolor{myrowcolor} 
\multirow{-3}{*}{\textsc{MIT-states}} \cellcolor{white} & \textsc{GDE (image)} & \bfseries 18.9 & \bfseries 34.2 & \bfseries 18.5 & \bfseries 25.3 & \bfseries 13.7 & \bfseries 3.3 & \bfseries 107.0 \%\\ 
\bottomrule 
\end{tabular}
}
    \caption{Compositional classification results of MERU's hyperbolic representations, closed-world scenario.}
    \label{tab:classification MERU (CW)}
\end{table}

\subsection{Runtime}
\newcommand{\bigO}{\mathcal{O}}
A potential limitation of our proposed framework is the additional computational costs it requires for mapping embeddings to and from the tangent space. We now analyze the inference time of the decomposition procedure. 

Suppose we compute a decomposable set for $M = |\Z|$ composite concepts using $N = |\T|$ visual embeddings on the sphere $\Sphere^{d-1}\subset \Real^d$.
Compared to LDE, GDE additionally computes $\Log_\mu$ for the $N$ inputs and $\Exp_\mu$ for the $M$ tangent compositions. The computational complexity of these operations is $\bigO(Nd)$ and $\bigO(Md)$, respectively. 
Note that the orthogonal projection in \cref{eq:log map sphere} can be rewritten as $(I_d- \mu\mu^\top)\mathbf{w} = \mathbf{w} - (\mu^\top\mathbf{w})\mu$, avoiding the explicit computation of the $d\short{\times}{1} d$ matrix.
Calculating $\mu$ with \cref{alg: intrinsic mean} is $\bigO(Nd)$ per gradient step, keeping the extra compute linear in $N, M, d$.
\cref{tab:computational costs decompositions} reports the runtimes for GDE, LDE, $\mu, \Log_\mu, \Exp_\mu$ in our experiments (tolerance for $\mu$ is $\epsilon\shorteq10^{-5}$).
Both methods are fast on the relatively small datasets used for our analysis. 
GDE is significantly slower than LDE, with most of its extra runtime being spent on the computation of $\mu$. However, we argue that approximating $\mu$ with a smaller subset of $N'<N$ input embeddings could be sufficient and drastically improve efficiency when the number of inputs is large.

\begin{table}[h]
    \centering
\resizebox{\linewidth}{!}{
\begingroup
\setlength{\tabcolsep}{2.5pt} %
\begin{tabular}{l cc cc cccc} 
   \toprule 
   Dataset & $N$ & $M$ & LDE & GDE & $\mu$ & steps $\mu$ & $\Log_\mu$ & $\Exp_\mu$ \\ 
   \hline 
    \textsc{UT-Zappos} & 22998 & 192 & 41$\pm$1 ms & 382$\pm$9 ms & 267$\pm$15 ms & 3 & 83$\pm$4 ms & 0$\pm$0 ms \\
    \textsc{MIT-states} & 30338 & 28175 & 149$\pm$0 ms & 850$\pm$8 ms & 560$\pm$15 ms & 5 & 106$\pm$4 ms & 60$\pm$1 ms \\
    \textsc{Waterbirds} & 4795 & 4 & 7$\pm$0 ms & 59$\pm$14 ms & 43$\pm$12 ms & 4 & 11$\pm$2 ms & 0$\pm$0 ms \\
    \textsc{CelebA} & 162770 & 4 & 375$\pm$4 ms & 3812$\pm$35 ms & 2902$\pm$37 ms & 5 & 557$\pm$14 ms & 0$\pm$0 ms \\
   \bottomrule
\end{tabular}
\endgroup
}
\vspace{-9pt}
    \caption{Runtimes on a Titan Xp GPU, averaged over 5 runs.}
    \label{tab:computational costs decompositions}
\end{table}
\vspace{-4pt}

\subsection{Generated images}
\label{sec:extra generated images}

\begin{figure*}
    \centering
    \includegraphics[width=\textwidth]{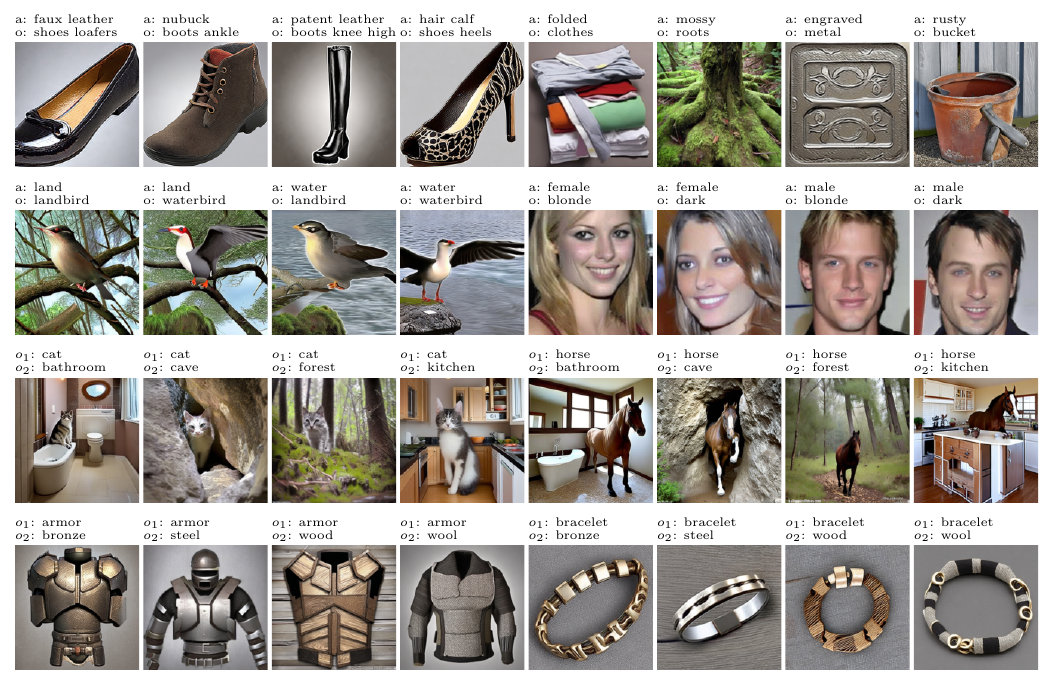}
    \caption{
    Additional generated images obtained by inverting the decomposable embeddings computed with GDE, using StableDiffusion with the unCLIP technique.
    We include attribute-object pairs from the UT-Zappos and MIT-states datasets (\textit{first row}), and from the Waterbirds and CelebA datasets (\textit{second row}). Similarly to the animal-animal pairs shown in the main document, we visualize animal-environment pairs (\textit{third row}) and item-material pairs (\textit{fourth row}) from the MIT-states objects.
    }
    \label{fig:SD extra}
\end{figure*}

In \cref{fig:SD extra} we show extra images generated using StabeDiffusion with the unCLIP module to invert composite embeddings. 
We include attribute-object pairs from all the datasets used in our experiments. Similarly to the animal-animal pairs shown in the main document, we identify other high-level categories within the MIT-states objects (items, environments and materials) and visualize animal-environment and item-material compositions.

\begin{figure*}[t]
    \centering
    \includegraphics[trim={0 10 0 0}, clip, width=\textwidth]{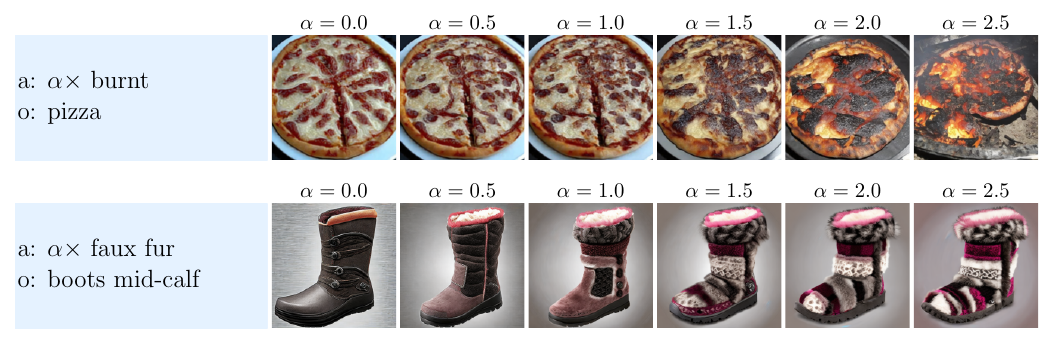}
    \caption{Scaling attribute direction in attribute-object compositions.}
    \label{fig:SD scaling attr}
\end{figure*}

Also, we observe that the modularity of the compositions allows to gain finer control on the composite embeddings. In \cref{fig:SD scaling attr}, we invert embeddings of the form $\Exp_\mu(\alpha\vv_a + \vv_o)$, where the attribute direction is scaled by a scalar $\alpha\in \Real$.
In the generated images, changing the value of $\alpha$ modifies the intensity of the attribute that results in a lower or strong appearance of it in the generated images. 
This experiment further demonstrates that the primitive vectors resulting from solving the proposed optimization problem are interpretable directions of the latent space.

Our initial goal of the generative experiments was to qualitative inspect the GDE compositions. However, the good quality of the results suggests that our framework could be useful for augmenting compositional datasets. To support this, we compute the average CLIP-score of 500 outputs (five generations of 100 random unseen concepts), to assess how a CLIP model perceives composite concepts in generated images. As a baseline, we consider the default text-to-image (T2I) version of the generative model.
\begin{table}[h]
    \centering
\resizebox{0.95\linewidth}{!}{
\begin{tabular}{cc c cc} 
   \toprule 
   \multicolumn{2}{c}{\textsc{UT-Zappos}} &&  \multicolumn{2}{c}{\textsc{MIT-states}} \\
   \midrule
   GDE: 0.68$\pm$0.06 & T2I: 0.62$\pm$0.10 && GDE: 0.58$\pm$0.08 & T2I: 0.55$\pm$0.10 \\ 
   \bottomrule
\end{tabular}
}
    \caption{Average CLIP-score of SD generated images.}
    \label{tab:SD CLIP-scores}
\end{table}

\subsection{Failure cases}
\begin{figure}[ht]
    \centering
    \includegraphics[width=.95\columnwidth]{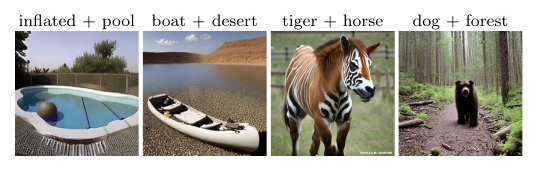}
    \vspace{-8pt}
    \caption{Failure instances in SD generations.}
    \label{fig:SD failures}
\end{figure}

We investigate failure cases in Stable Diffusion visualizations and noted that the decomposable embeddings may encode spurious correlations of the input data or produce ambiguous compositions. 
For instance, the generated images in \cref{fig:SD failures} suggest that the 'inflated' and 'boat’ primitive directions are respectively linked to 'round object' and 'water’, and the 'tiger’+'horse' and 'dog'+'forest' compositions are respectively close to 'zebra' and 'bear'.

\vspace{-5pt}

\end{document}